\pdfoutput=1

\documentclass[11pt]{article}

\usepackage[]{naacl2021}

\usepackage{times}
\usepackage{latexsym}

\usepackage[T1]{fontenc}
\usepackage{multirow,booktabs,colortbl,tabularx}

\usepackage[utf8]{inputenc}
\usepackage{graphicx}
\usepackage{tikz}
\usepackage{amsmath}
\DeclareMathOperator*{\argmax}{argmax}
\usepackage{enumitem}
\usepackage{subcaption}
\usepackage{tablefootnote}
\usepackage{microtype}
\usepackage{listings}

\usepackage{booktabs, makecell, tabularx}
\newcolumntype{L}{>{\raggedright\arraybackslash}X}
\usepackage{siunitx}
\usepackage{caption,afterpage}
\title{Interpretation of Black Box NLP Models: A Survey}

 \author{Shivani Choudhary, Niladri Chatterjee, Subir Kumar Saha \\
  \texttt{\{shivani@sire, niladri@maths, saha@mech\}.iitd.ac.in}\\
   Indian Institute of Technology Delhi \\
  Hauz Khas, Delhi-110016, India\\}

\begin{document}
\maketitle
\begin{abstract}
Machine learning (ML) models are ubiquitous in every wake of life and employed in high-stakes decision-making tasks. The complexity involved in ML models, specifically in Neural models, has brought in the question, \textit{``How does a model make a decision?"}. An insight into the model's decision-making process will help fix the accountability for the model's decision. It will further address ethical, safety, and bias issues associated with the model. To address these interpretability can provide an explanation that humans can easily understand. This survey will present a comprehensive survey of methods employed for interpretability. Firstly, it presents the definition of \textit{Interpretability} followed by a discussion on the different approaches adopted in NLP space. Finally, It will highlight the synergy between different approaches and issues with interpretability in NLP space. 
\end{abstract}

\section{Introduction}
Machine learning (ML) is now ubiquitous in the current era. Some of the ML models, especially deep learning-based models, achieve near-human accuracy. It has led to the adoption of ML into several areas, including critical areas like health, financial markets, criminal justice. \citep{DBLP:journals/corr/Lipton16a}. It had raised an important question regarding the black-box nature of the ML models. 

There is no concrete definition of the Interpretability of ML. However, different authors have tried to provide some definitions. Interpretability is often considered along with explainability; even some of the authors have used it interchangeably \cite{DBLP:journals/corr/abs-2012-14261}. In a general sense, interpreting data means extracting information from them \citep{Murdoch22071}. It can be defined as \textit{``the degree of understanding of a model in regards to how a decision was made so that a model can provide an answer to a user''}. It can also be understood as the extraction of the relevant knowledge from the ML models that was \textit{``learned by the model''} or was \textit{``present in the data''} \citep{Murdoch22071}. It highlights the key bit of interpretability, a user should understand and reason the model output. Learned understanding can be presented in different forms like visualization, decision tree, natural language and graph as well. Interpretability is not limited to model parameters, learning algorithm, feature selection, or a combination of these \citep{8397411, DoshiVelez2017TowardsAR}. \citep{DBLP:journals/corr/Lipton16a} has listed \textit{trust, causality, transferability and informativeness} as key elements when considering the interpretability research.  

Most of the works in this area were focused on model working, on a global level or on a local level. However, only a handful of work tried to analyze the aspect of interpretability from a ``user'' prospect. In other words, they were answering the question ``Interpretable to Whom?'' because different levels of audience will interpret the model separately. E.g., A person with a Statistics major can interpret the behavior of Bayesian models to some extent, while a person who is not a domain expert may not find the exact information relevant for interpretation. So, It gives another dimension to interpretability-related research. 

Machine learning (ML) models are sensitive to the perturbation in the input. Sometimes, even a minor change can lead to a change in prediction, let alone the confidence of the prediction. This behavior of ML models can also help to get an explanation for a prediction. In the case of computer vision, it can help understand which pixels or superpixels can lead to a behavior change. 

This survey aims to present the theoretical overview of interpretability along with different interpretability methods. In section -- \ref{back}, we present an overview of the need for interpretability. Section -- \ref{IC} lists the classification of interpretability. Section -- \ref{IM} lists different methods of interpretability with their application in the NLP task. This section also lists some of the debate surrounding those models. Finally, Section -- \ref{obs} presents some of the findings with respect to different models followed by Section -- \ref{conc} as conclusion. Table -- \ref{table1} presents the categorization of models with a list of representative papers with the area of application in NLP.

\section{Background}
\label{back}
\subsection{Why do we need interpretable models?}
With broader adoption of the ML models in the various day-to-day interaction, it has brought in the aspect where a human needs to understand the behavior of ML models—it necessities a person to understand the process by which it has reached a particular conclusion. Engagement of ML models in activities like criminal recidivism, loan approval, premium calculation, etc., has brought in the ethical and fairness concerns in the ML adoption in real life. In order to understand or allay the apprehensions raised on the ML, there needs a requirement to understand the decision of the models.  

\subsubsection{Reliability}
Adoption of models in different areas does not require it to be reliable every time. But, the scenarios where a decision outcome from the models can have a big impact need to be reliable. In case of medical diagnosis, like detection of \textit{malignant} and \textit{benign} tumour. By just changing a small set of pixels can lead to an altered output from a trained DL network \citep{DBLP:journals/corr/abs-1804-05296}. Another adversarial example \citep{DBLP:journals/corr/abs-1807-03571} that leads to an incorrect prediction from the network from red light to green light. These predictions can cost a human life. So in these cases, reliability is required. Had the models been interpretable, the model's decision could have been explained. Specially, in case of an adversarial attacks \citep{DBLP:journals/corr/abs-1712-06751}.  

In another scenario, Husky vs. Wolf classifier, the models seem to have learned the snow patch to make the classification \citep{10.1145/2939672.2939778}. It has nothing to do with the features of the two breeds. In this case, due to some bit to interpretability of the decision outcome, we can understand the decision-making process. Though the predicted classes are incorrect, we know that this classifier has learned some irrelevant patterns and is not reliable. 

\subsubsection{Ethical concerns}
In recent times, it has been brought to focus that ML models are biased due to bias ingrained in the dataset or due to algorithmic complexity. ProPublica has presented an analysis where it has shown that COMPASS, which predicts criminal recidivism, has a bias towards native African American. In another case, Amazon's one-day delivery was unavailable to the minority neighborhood while it was available in other neighborhoods. All these concerns stress on the fact that we need interpretable models through which we can understand the decision outcomes. Whether the decision was made using protected attributes like gender, place of birth, caste, etc. 

\subsubsection{Research aspects}
Ml models learn different patterns from the data. The learned rules are a way to understand the unknown. Interpretability of the models will help us understand those undefined rules to understand the missing elements in fields like Physics, Genomics, etc. 

\subsubsection{Transparency}
ML models are treated as a ``black-box model''. Apart from those parameters and associated weights increase the model's complexity. A decision outcome generated from those models is hard to comprehend. With an interpretable model, we can easily understand the working and evaluate whether the model characteristics like causality, transferability, and informativeness \citep{DBLP:journals/corr/Lipton16a}. It will ensure that we would be able to evaluate the working of models in the case of real-world data. Then we can answer the questions from causality prospects like ``what if?'', ``why?''. It will enable a user to evaluate the model from counterfactual criteria. 

\section{Interpretability classification}
\label{IC}
Interpretability of a model can be addressed by several approaches depending upon the type of end explanation it generates. Explanation methods can be categorized based upon different scopes and properties. 
\subsection{Local vs. Global interpretability}
Interpretability in this context can be understood as what part of the model can be interpreted. How much insight the generated explanation can provide.

\begin{figure*}
\begin{subfigure}{.5\textwidth}
    \centering
    \includegraphics[width=0.9\textwidth]{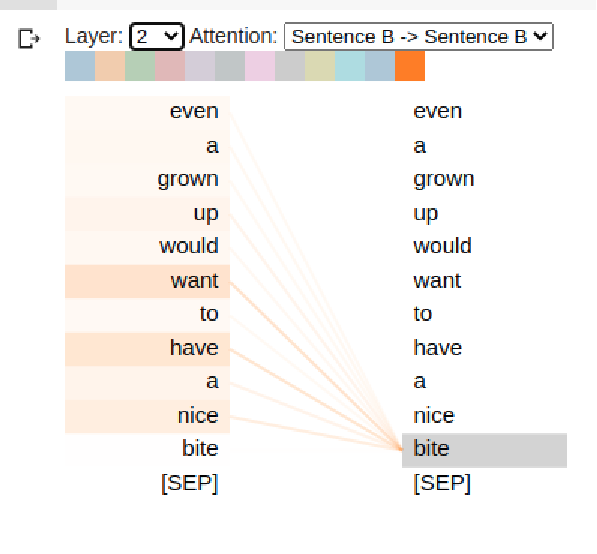}
    \caption{Attention from Sentence B to B, generated using \citep{vig-2019-multiscale}}
    \label{fig:Self-Att}
\end{subfigure}
\begin{subfigure}{.5\textwidth}
    \centering
    \includegraphics[width=0.9\textwidth]{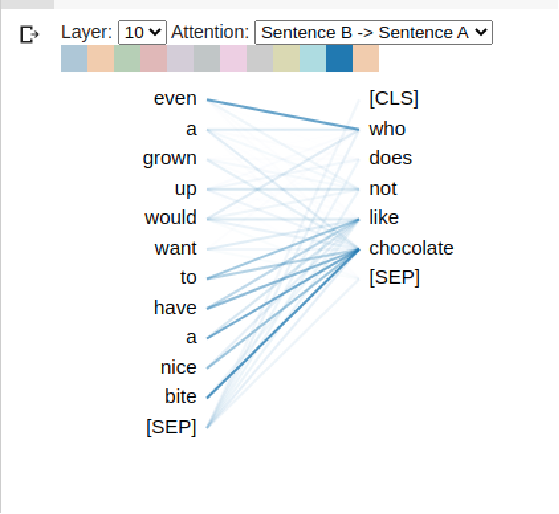}
    \caption{Attention from Sentence B to A, generated using \citep{vig-2019-multiscale}}
    \label{fig:Cross_Att}
\end{subfigure}
\end{figure*}
\subsubsection{Global Interpretability} 
A model can be categorized as globally interpretable if we can comprehend a model's behavior altogether from the parameter, weights, and other ancillary information \cite{DBLP:journals/corr/Lipton16a}. It demands the understanding to the extent to which a user can figure out the interplay between features and the weightage of features. Global interpretability centers on how well parametric variation and association can be comprehended by humans. In real-life models, there are a huge number of parameters associated with the algorithms. The complexity of the global interpretation is added by the number of features present in the data. With a lot of features in hand, it is difficult to interpret a model globally \citep{Honegger2018SheddingLO,doi:10.1177/0963721409359277}. In such cases, we need to employ algorithms like t-SNE to present higher dimension data points in a lower-dimensional space \citep{Maaten2008VisualizingDU}. 

In this class of models are decision tree, linear regression, rule extractor, where we can either interpret the rule from plain text, by the split at the nodes or by examining the weights. 

\subsubsection{Local Interpretability}
Local interpretability is more focused on understanding the system's behavior concerning a single data sample or a group of samples in the neighborhood. A complex model is usually comprehended by an approximate linear model. A good approximation can be achieved with a sacrifice of accuracy for this model. In this case, the model's behavior can be analyzed for a single case or a group case. When a group of data points is considered then, approximate will be performed the local interpretation task for each of the samples \citep{molnar2019}. 

\subsection{Inherently interpretable vs Post-hoc explanation}
Models can be broadly categorized in the Inherently interpretable or Post-hoc explanation-based interpretable model \citep{DoshiVelez2017TowardsAR}. Inherently interpretable models are those which are by design interpretable. Due to constraints, these models are often mentioned as ``white-box models" \citep{Rudin2018PleaseSE}. These models are easy to interpret. As  \citep{jacovi-goldberg-2020-towards} has pointed out that a model may not be interpretable just by claiming it to be interpretable. It needs to be verified, as is the case with \textit{Attention mechanism} in NLP.

Interpretation via post-hoc techniques is applied after the model is trained. The weights are interpreted using different local and global analysis techniques. Post-hoc explanation can be applied to the inherently interpretable models as well. 

In the above two interpretations, we are concerned with the model's behavior. Another aspect of interpretability can be driven from the data itself, even before it is consumed into the ML models/algorithm. Some definitive trends can be extracted from the data by doing exploratory analysis. This pre-training information helps to interpret the models.

\subsection{Model specific and model agnostic}
Post-hoc explanations can be categorized into two parts based on their application and relevance to the models' internals. Model-specific methods are dependent on the internals of the model. These types of methods can be applied to a specific family of models. On the other hand, model agnostic methods can be applied to any model. Model agnostic methods consider any ML models as black box. So it does not have any information about the internal organization of the model \citep{molnar2019}. 

\subsection{NLP Specific}
Interpretability in natural language processing (NLP) is an important aspect of understanding why a response to a specific question was made. We can take sentiment analysis as an example. Under this, a model classifies a sentence as positive or negative sentiment based upon the sentence. In order to explain the outcome, we need to understand different features of a sentence like tokens (word may be in original form or base form), syntactical structure, punctuation, etc. 

The analysis would require finding the token that has carried more weightage than others in the decision-making process. BERT \cite{Devlin2019BERTPO} is based on \textit{Attention mechanism} which is considered as inherently interpretable in nature. A visualization of attention between two sentences from different layers gives an insight into how each word influences each word. As an example, two sentences are taken as input \textit{"Who does not like chocolate"} and \textit{"Even a grown-up would want to have a nice bite"}. Figure-\ref{fig:Self-Att} and Figure-\ref{fig:Cross_Att} shows the different level of influence between the words.   
Interpretability of the models would provide different types of explanation. Visualization, a summary of feature importance are one of those examples. 

In order to address the interpretability, some aspects that can be probed will be explained in the survey are listed below. These are further categorized into two parts Local and Global explanation. 
\begin{itemize}
    \item Local Explanation
    \begin{description}
            \item [Feature based:] These methods focus on the task; what are the essential features that have impacted the model's decision outcome? It is further divided into different methods based on the type of model's feature/data point it uses to generate an explanation. 
            \item [Causality-based scenario:] How would model scenario when it is tested from the Adversarial and counterfactual viewpoint? It evaluates the model robustness and performance in an alternate scenario. 
            \item [Natural Language Explanations:] This approach generates explanation that can be understood by a layman. 
    \end{description}
    \item Global Explanation
    \begin{description}
            \item[Visualization:] How does each word influence each other in the model's definition?
            \item[Probing:] It tries to analyse and evaluate, ``what are the linguistic features that are captured by the model?"
    \end{description}
\end{itemize}

\section{Interpretability Methods}
\label{IM}
\subsection{Feature based}
\label{feature}
\subsubsection{Gradient based methods}
\label{gradient}
Simonyan et al. \citep{Simonyan2014DeepIC} has proposed using the gradient of the output with respect to pixels of an input image to compute a ``saliency map" of the image in the context of image classification tasks. In NLP domain, it is transformed into taking the gradient of output logits with respect to input. It measures the effect of change in input to the output generated by the model. Common methods based on the use of gradients are DeepLift \citep{Shrikumar2017LearningIF}, Layerwise relevance propogation (LRP) \citep{Binder2016LayerWiseRP}, Guided-back propagation \citep{Springenberg2015StrivingFS}, deconvolutional networks \citep{Zeiler2010DeconvolutionalN}. Gradient extraction helps to identify the important features for a given prediction. These method faces issue with sensitivity and implementation invariance. It means if two inputs with one differentiating feature (token) leads to a change in prediction then it needs to treated as an important feature. Sundararajan et.al. \citep{10.5555/3305890.3306024} proposed Integrated gradient (IG) method which can address the issue of sensitivity and implementation invariance. IG works on the approach where gradients are accumulated for all the points on a straight line between an input and a baseline point. He et. al. \citep{he-etal-2019-towards} has applied this approach in neural machine translation (NMT) task and Mudrakarta \citep{mudrakarta-etal-2018-model} applied it in Question-Answering task to understand the keywords which were influencing the answers. Arras et.al. \citep{DBLP:journals/corr/ArrasHMMS16} has used LRP to analyse CNN trained for topic categorization task.  Wang et. al. \citep{Wang2020GradientbasedAO} has shown that gradient based analysis can be manipulated. 

\begin{table*}[]

    \sisetup{group-minimum-digits=4,
             group-separator={,},
             }
    \small
    \setlength\tabcolsep{4pt}
\begin{tabularx}{\linewidth}{@{} l l c
                             L
                             c}
    \toprule
Methods    
    &   Approach 
        &   Type of Analysis    
            &   {\makecell{Representative Paper}}
                &   {\makecell{NLP domain \\Application}}\\

    \midrule
\multicolumn{5}{c}{Local Interpretability}                                                                                  \\ \midrule
\multirow{4}{*}{Feature based} 
    &   \makecell{Gradient based}        &  Model agnostic   
            &   \makecell{\citep{Shrikumar2017LearningIF}\\
            \citep{Binder2016LayerWiseRP}\\
            \citep{Springenberg2015StrivingFS}\\
            \citep{Zeiler2010DeconvolutionalN}} 
                &   \makecell{NMT, QA, \\Topic classification} \\
                \addlinespace
    & \makecell{Input\\ perturbation}         &  Model agnostic 
            &   \makecell{\citep{Ribeiro2016WhySI}\\
            \citep{Ribeiro2018AnchorsHM}\\
            \citep{rychalska-etal-2018-much}} 
                &    \makecell{QA} \\
                \addlinespace
    & \makecell{SHAP}         &   Model agnostic   
            &   \makecell{\citep{Lundberg2017AUA}\\
            \citep{Lundberg2018ConsistentIF}\\
            } 
                &   \makecell{QA, \\Text classification} \\
                \addlinespace
    & \makecell{Attention based}         &   \makecell{Inherently interpretable\\ and model specific\footnotemark}
            &   \makecell{\citep{Tu_Huang_Wang_Huang_He_Zhou_2020}\\
            \citep{10.5555/3157096.3157129}\\
            \citep{Li2020TowardIO}\\
            \citep{ijcai2019-714}} 
                &   \makecell{QA, VQA \\Sentiment Analysis } \\
                \addlinespace
\midrule
\multirow{2}{*}{Causality based} 
    &   \makecell{Adversarial \\examples}        &   Model Specific   
            &   \makecell{\citep{Ebrahimi2018}\\
            \citep{Ribeiro2018a}\\
            \citep{Sato2018b} } 
                &   \makecell{NMT, VQA, \\Sentiment Analysis,\\
                Grammatical error\\ detection} \\
                \addlinespace
    & \makecell{Counterfactual\\ explanation}         &   \makecell{Model agnostic \\ and\\ Model specific}   
            &   \makecell{\citep{Wu2021}\\
            \citep{Ross2021}\\
            \citep{Raffel2020ExploringTL}\\
            \citep{Elazar2021a}\\
            \citep{Vig2020}\\
            \citep{Finlayson2021}} 
                &    \makecell{Bias in model,\\ Syntactic evaluation\\ POS
                } \\
                \addlinespace
\midrule
NLE
    &   \makecell{Natural language\\ explanation}       &   Model Specific   
            &   \makecell{\citep{Park}\\
            \citep{Ling2017a}\\
            \citep{Tim2018}\\
            \citep{Kumar2020a}\\
            \citep{Mccann2019}} 
                &   \makecell{NMT,\\Label prediction, \\Natural language\\ inference\\}\\
                \addlinespace
\midrule
\multicolumn{5}{c}{Global Interpretability}                                                                                  \\ 
\midrule
Visualization
    &   \makecell{Visualization}       &   Model Agnostic   
            &   \makecell{\citep{Park2017RotatedWV}\\
            \citep{Li2016}\\
            \citep{Shin2018InterpretingWE}} 
                &   \makecell{Linguistic features}\\
                \addlinespace
\midrule
\multirow{2}{*}{Probing} 
    &   \makecell{Distributional word \\embedding probing} &      
            &   \makecell{\citep{Ebrahimi2018}\\
            \citep{Ribeiro2018a}\\
            \citep{Sato2018b} } 
                &   \makecell{NMT, VQA, \\Sentiment Analysis,\\
                Grammatical error\\ detection} \\
                \addlinespace
    & \makecell{Hidden state\\ probing}         &   \makecell{Model agnostic \\ and\\ Model specific} 
            &   \makecell{\citep{Shi2016a}\\
            \citep{Belinkov2017a}\\
            \citep{Mareek2020AreMN} \\
            \citep{Conneau2018}\\
            \citep{Hupkes2018}\\
            \citep{Peters2020}} 
                &    \makecell{NMT,\\ Compositionality,\\ Correference resolution \\syntactic feature
                } \\
                \addlinespace            
\bottomrule
\end{tabularx}
\caption{List of interpretability methods in NLP. This table is separated in two parts -- global method or local method. \\
\emph{Note-1: }Method column presents the broader classification of methods. Second column presents the fine categorization under the broader category. Last column lists down NLP task employed to interpret the model's behaviour. Type of Analysis is an approximate categorization by taking the features used by Interpretability method. \\
\emph{Note-2: }\textit{It is not an one-to-one mapping with the representative paper.Details of each broader classification is presented in section-\ref{IM}. List of papers is not exhaustive. There are some of the papers that are related to different methods are listed in the discussion}.
}
\afterpage{\footnotetext{A detail discussion whether Attention is inherently interpretable or not}}
\label{table1}
\end{table*}

\subsubsection{Input Perturbation Based}
\label{inputpertu}
It is another method to extract the importance of the different features present in the input samples. In this method, a word (token) or a collection of words (tokens) are modified or removed from the input samples, and a resulting change is measured. Feature importance is measured by the drop in the performance of the model. If the drop is high then the feature is very important for the model. These methods are model agnostic in nature. 

Ribeiro et.al. \citep{Ribeiro2016WhySI} proposed a locally interpretable and model agnostic explanation (LIME) framework. In this method the model under consideration is assumed as a black box model. The central idea of LIME is to generate a local surrogate, a glass-box model, to generate explanations for the decision outcomes. LIME generates a dataset with perturbed inputs and corresponding predictions from the black box model.  On this new dataset, LIME trains an interpretable model, which is weighted by the proximity of the sampled instances to the instance of interest. Mathematical formulation of LIME as below
\begin{equation} \label{eqn}
  \xi(x) = \argmax_{g \in \mathcal{G}} \mathcal{L}(f, g, \pi_x) + \Omega(g)
\end{equation}

\begin{itemize}[noitemsep]
    \item f(x) is the prediction from the black model for sample x
    \item $\mathcal{G}$ is the class of potential interpretable models
    \item $\pi_x$ defines the size of the neighbourhood
    \item $\mathcal{L}$ is the Loss that will measure the closeness of explanation
    \item $\Omega(g)$ determines the complexity of the local surrogate models
\end{itemize}

In original LIME method the analysis is based on the word level (single token), later they proposed a new model that is based on consecutive tokens \citep{Ribeiro2018AnchorsHM}. Consecutive tokens are called as \textit{Anchors}. Anchors explains individual predictions of any black-box classification model by finding a decision rule that ``anchors” the prediction sufficiently. A rule anchors a prediction if changes in other text does not change the prediction.
LIME out can vary significantly even if two artificial points are in proximity \citep{AlvarezMelis2018OnTR} and Slack et. al. pointed out that it is prone to adversarial attacks \citep{10.1145/3375627.3375830, Tan2019WhySY}.  Different version of LIME are proposed Zafar et. al. proposed D-LIME \citep{Zafar2019DLIMEAD}, Zhou et.al. proposed S-LIME \citep{10.1145/3447548.3467274}.

LIME has been employed in QA task by Basaj et.al. \citep{rychalska-etal-2018-much} to check how many words from question are relevant to predict correct answer. Sydorova et.al. has applied it for QA task in conjugation with knowledge base\citep{Sydorova2019InterpretableQA}. 

\subsubsection{SHAP}
\label{SHAP}
SHAP (\textit{SH}apley \textit{A}dditive ex\textit{P}lanations) was proposed by Lundberg and Lee \citep{Lundberg2017AUA}. It is based on game theory based Shapely Values \citep{shapley1953value}. Methods like LIME may not distribute attributions fairly among the features while Shapely value guarantees it \citep{molnar2019}. A way to use the efficient distribution using Shapely value would be to compute shapely values for each and every combination of the features ( a power set of the features)  by training a linear model. But, it will be computationally expensive to train $2^M$ models for M set of features. 

SHAP calculates Shapely value and presents it as a linear model or additive feature attribution. SHAP presents a model explanation as 
\begin{equation} \label{eqn2}
    g(x')={\phi_0 + \sum_{j=1}^{\mathcal{M}}\phi_jz_j'}
\end{equation}

Where $g$ is an explanation model, $\mathcal{M}$ is the maximum size of coalition, $\phi_j$ is the feature attribution for feature $j$  and $z'$ is the binary vector.

The model agnostic version of the SHAP is Kernel SHAP. Lundberg and Lee call it as LIME + Shapely values. The solution of equation \ref{eqn} will satisfy the property of local accuracy, missingness and consistency \citep{Lundberg2017AUA}. Finding those values heuristically would be problematic. They suggested following would be choice for the parameters in equation \ref{eqn}.
\[
    \Omega(g) = 0 \]
\[
    \pi_x(z') = \frac{(M-1)}{(M choose \vert z' \vert)\vert z' \vert(M - \vert z' \vert)}\\
\]
\[
    \mathcal{L}(f, g, \pi_x') = \sum_{(z' \in Z)} [f(h_x(z') - g(z')]^2 \pi_x(z')
\]

where $\vert z' \vert$ is the number of non-zero elements in $z'$, $h_x(z)$ is the mapping function that maps the combination $z'$ to original feature space. Deep-SHAP (DeepLIFT + Shapely values) is a model specific version of SHAP. Lundberg et. al. \citep{Lundberg2018ConsistentIF} proposed TreeSHAP for tree based machine learning models. TreeSHAP has issues with giving importance to the non-important features as well. Some of the authors have also highlighted that SHAP is prone to adversarial attack \citep{10.1145/3375627.3375830}.

In NLP, Zhao et. al. \citep{Zhao2020SHAPVF} has developed SHAP to explain CNN based text classification model. Balouchzahi et. al. \citep{Balouchzahi2021FakeNS} has used it for fake news profiling using SHAP based feature selection. Wu et. al.\citep{Wu2021PolyjuiceGC} has used it to generate counterfactual.

\subsubsection{Attention based}
\label{attention}
Attention mechanism was proposed by Bahdanau et. al. \citep{Bahdanau2015NeuralMT}. Attention is a weighted sum of the intermediate representation in neural network. Attention weights from the attention based models can be used for local interpretation. Attention mechanism has gained traction in NLP task. It is a state-of-art architecture for the NLP task likes question answering, Neural machine translation, Visual question answering etc. In NLP context, the feature (token) with higher weight is considered as an important feature. Attention has been applied to question answering task \citep{Tu_Huang_Wang_Huang_He_Zhou_2020, sydorova-etal-2019-interpretable, 10.1145/3209978.3210081}, dialogue suggestion system \citep{Li2020TowardIO} and sentiment analysis \citep{ijcai2019-714, Yan2021SAKGBERTEL, ijcai2018-590}. This method has been applied to multimodal data \citep{10.5555/3157096.3157129} like visual question answering. It uses both text and image mode of data as input. Lu et. al. \citep{10.5555/3157096.3157129} called it \textit{co-attention} where the reasoning is performed with question attention and visual attention. Combining the attention weights with visualization helps to interpret the model. 

Several authors has used \textit{Attention} in different NLP task. But there is an ongoing debate ``\textit{Is attention interpretable}" \citep{Pruthi2020LearningTD, Serrano2019IsAI, Jain2019AttentionIN, wiegreffe-pinter-2019-attention, Vashishth2019AttentionIA}. \citep{Jain2019AttentionIN} and \citep{Vashishth2019AttentionIA} has presented two arguments.
\begin{enumerate}
    \item Attention weight should correlate with feature importance similar to gradient based methods
    \item Alternative attention weights (counterfactual) should lead to changes in the prediction
\end{enumerate}

Both the premise were not fulfilled in their experiments on question answering task and Natural language inference task. on the other hand \citep{Serrano2019IsAI} had found that alternative weights did not necessarily resulted in outcome change. However, these arguments were countered by \citep{wiegreffe-pinter-2019-attention} and argued that model's weight are learned in a unified manner with their parameters. So, detaching attention score from parts of the model will degrade the model itself. They also argued that \textit{Attention} is not the only explanation. \citep{Vashishth2019AttentionIA} has performed experiments on tasks like text classification, Natural language inference (NLI) and NMT, and concluded that the model's performance is dependent on the type of task. Attention weights are interpretable and correlate with feature importance --- when weights are computed using two sequence which are the function of input and Attention weights may not be interpretable when the score is calculated on single sequence like Text classification. 

\subsection{Visualization}
\label{visualization}
Visualization is an important way to understand how a neural model work \citep{Li2016}. It can be applied with any of the feature importance based methods. With visualization, we can project the feature importance weights using heatmap, partial dependency plot etc. Most of the state-of-art NLP task are dependent on the word embedding. Sparse encoding like one-hot encoding has been replace by dense encoding like (word2vec\citep{Mikolov2013EfficientEO}, Glove \citep{Pennington2014GloVeGV} and representation from intermediate layers of BERT \citep{Devlin2019BERTPO} and ElMO \citep{Peters2018DeepCW}). Word embedding based information captures information at model level. Hence it presents the information at global level. It presents which type of linguistic features are learnt by the model. Dense word embedding is presented in hyper-space. In order to understand the embedding, it needs to projected into two or three dimensional space.
\begin{figure}[!htb]
    \includegraphics[width=0.4\textwidth]{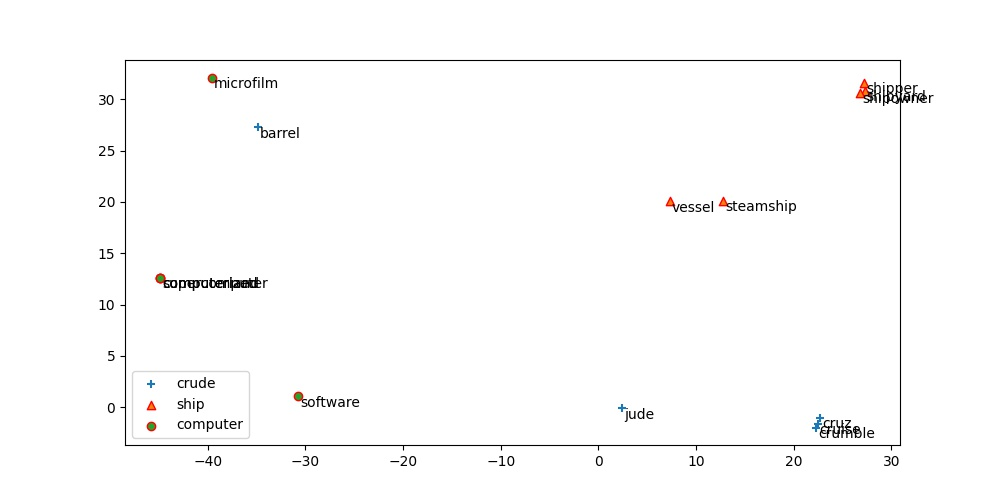}
    \caption{t-SNE 2D projection of FastText embedding \footnotemark trained for 50 epoch on Reuters news corpus from NLTK, with context len 15}
    \label{fig:fig3}
\end{figure}
\footnotetext{\url{https://radimrehurek.com/gensim/auto\_examples/tutorials/run\_fasttext.html}}
t-SNE \citep{Maaten2008VisualizingDU} and principal component analysis are two important tools to present a high dimensional representation to a lower dimensional space. \citep{Li2016} has presented how individual components get activated by adding negative and positive words to a sentence. Using t-SNE, they have also shown that neural model learn the properties of local compositionality, clustering negation+positive words (`not nice’, `not good’) together with negative words. \citep{Park2017RotatedWV} has suggested that rotation of the embedding can lead to an increased intepretability. His method is based on exploratory factor analysis and used PCA to visualize the representation. \citep{Shin2018InterpretingWE} has presented Eigen vector based method to analyse word embedding. An example of t-sne projection of FastText \citep{Bojanowski2017EnrichingWV} embedding of 5 nearest words to the words crude, ship and supercomputers in Figure-\ref{fig:fig3}.

\subsection{Probing}
With dependence of state-of-art NLP models on dense vector representation of the words \citep{Devlin2019BERTPO, Mikolov2013EfficientEO, Peters2018DeepCW}, it is pertinent to ask question like which type of linguistic features are encoded in the Word embedding or in the intermediate representation of the Neural models \citep{Rogers2020a,Zhang2019e, Poliak2020a}. A method to extract these information is called as probing. Probing tasks are commonly known as ``auxillary tasks" \citep{Adi2017}, ``diagnostic classifier" \citep{giulianelli-etal-2018-hood, Hupkes2018} or decoding. Under this task, an external classifier is trained on the intermediate representation or word embedding to predict the liguistic property under the observation. It provides an insight into the fact that how well a model has learned the specified linguistic property. Several linguistic features have been analysed to extract different properties like Morphological, syntactic and semantic\citep{Voita2020, Tang2021}. It is based on the premise that if there is more task relevant information is learned by the model then it is going to perform better on the presented task. However, some researchers have pointed out that probe selection and measurement should be carefully done in order to get a reliable insight into the model \citep{Voita2020, Hewitt2019}. Linguistic insight can be extracted from the representations from intermediate layer of a neural model and word embedding. In this survey, we divide the probing into two parts -- Hidden state probes and Distributional word Embedding probes. 
\subsubsection{Distributional word Embedding probes}
\label{Eprobing}
Early word embedding methods were based on the distributional hypothesis, meaning information captured by a word can be extracted by the negihbourhood in which it was present. Two notable distributional word embedding CBOW/Skip-gram \citep{Mikolov2013EfficientEO} and GloVe \citep{Pennington2014GloVeGV}. Several author has tried to extract the information contained in the word embedding using simple classifiers like -- logistic or linear classifiers. They all have reported the presence of the linguistic concepts to a varied extent \citep{Kohn2016a, Utsumi2020, Rubinstein2015, Gupta2015}. \citep{ghannay-etal-2016-word} has further extended the probing using neural network for POS tagging, Named entity recognition and Mention detection. Apart from word embedding, sentence embedding based analysis has also been attempted  by \citep{Adi2017} using LSTM. They have noted the effectiveness of CBOW in LSTM based encoder task. \cite{Conneau2018} and \citep{Tenney2019a} has presented a exhaustive list of task for the evaluation. This list was further extended by \citep{sorodoc-etal-2020-probing,Sahin2020}.

\subsubsection{Hidden state probing}
\label{Hprobing}
Distributional embedding has been replaced by contextual embedding like BERT \citep{Devlin2019BERTPO}, ElMO\citep{Peters2018DeepCW}  for NLP task. These embedding represent words as representations learned from the hidden states. Before we presents the work that examined deep-contextual embedding, we would like to highlight the works that has analysed the hidden state learned by neural models for different task. \citep{Shi2016a} has used probing technique in NMT task to determine whether LSTM based encode-decoder architecture can learn the syntactic features. They employed logistic classifier (as a diagnostic classifier) to predict different syntactic labels on top a learned sentence encoding vector and word by word hidden vectors. They pointed out that LSTM based encoder-decoder was able to learn different syntactic feature from the input sentence and different layers learned different features. They probed the model for 5 features 3 sentence level features --- Voice, Tense and Top level syntactic sequence and 2 word level features --- Parts of speech and smallest phrase constituent. \citep{Belinkov2017a} has extended the work of Shi et. al. for NMT task.\citep{Raganato2018} has used probing for attention based models and \citep{Mareek2020AreMN} has extended for multi-lingual task based on the \citep{Conneau2018} 10 linguistic task for probing. 

Use of probing is not limited to NMT task only. \citep{Hupkes2018} has used it to find the learning capabilities of neural model in the context of hierarchical and compositional semantics. This work was performed on artificial task to solve arthematic problem solving. \citep{giulianelli-etal-2018-hood} has used probing on subject verb agreement task by probing LSTM layers. \citep{Cohen2018} has applied probing in information retrieval. Probing has been profusely applied to analyse the learning capabilities of deep-contextualized embedding. \citep{Lin2019a,Clark2019,Tenney2019a, Yu2020, Peters2020} has applied probing to analyse different linguistic features. \citep{Lin2019a} has applied it to analyse syntactical feature. \citep{Clark2019} has applied probing on top of BERT's attention weights to analyse syntactic relation. They found that BERT can able to learn syntactical features, even if it is trained in unsupervised manner. \citep{Peters2020} applied probing on ELMO embedding and showed it can learn hierarchy of contextual information like lower layer representation performed better in POS tagging task and higher layer in correference resolution. In recent days, some of the new publication has applied it to behavioural explanation, phrasal representation and composition, conversational recommendation and to check the understanding of idioms \citep{Yu2020, Elazar2021, Tan2021, Penha2020}.

With the pervasive use of Probing in NLP, there is a word of caution came from \citep{Hewitt2020, Belinkov2021}. \citep{Hewitt2020} has pointed out that a good score on a particular NLP task may not provide the true picture. The performance may be due to the learning capabilities of the probe itself. Such problems were acknowledged by \citep{Zhang2019e} but a comprehensive analysis was put forth by Hewitt et. al. They have stressed on the fact that a good probe should have a good selectivity. To overcome the shortcomings, a concept of control task is proposed. Other authors have extended this concept in their works \citep{ravichander-etal-2021-probing, pimentel-etal-2020-information}.

\subsection{Natural language explanation}
\label{NLE}
Methods which are presented in this survey or otherwise is not suitable for a layman. It is meant to be used by ML practitioners. So, it is imperative to use methods that can generate explanations for a layman person. It means the explanation generated by interpretability methods can be presented in a simple language or may be as a summary. Natural language explanation (NLE) has already been applied in the computer vision domain by . It was applied for the task like self-driving cars\citep{Kim}, visual question answering task \citep{Park} and algebraic equation solving task by \citep{Ling2017a}. This method has been applied in the NLP area as well. \citep{Tim2018} proposed a two step process a two step process --- ``explain first then predict (reasoning)" and ``predict first then explain (rationalization)". They argued that ``explain first then predict" is more intuitive than the latter one. \citep{Kumar2020a} has proposed a model called NILE which follows the ``explain first then predict" strategy to derive NLE. NILE generates multiple explantion, one for each label, the predict the answer based on the explanation. \citep{Mccann2019} has proposed a model called as CAGE to generate an explanation for commonsense question answering. Language model inside the CAGE is based on GPT-2, a transformer \citep{Vaswani2017AttentionNeed} based architecture. NLE methods that are presented in this survey broadly falls in the local explanation category. These methods generates an explanation for a single instance. Model proposed by \citep{Tim2018,Kim, Ling2017a, Mccann2019} falls in post-hoc category because the explanation is generated after the model is trained. However, NILE can be categorized to inherently interpretable method because it first generates the explanation. 

\subsection{Counterfactual explanations (CF) and Adversarial examples (AE)}
\label{CEAE}
Machine learning models tries to learn the correlation between the features and labels. Any statistical correlation is acceptable in ML framework, without considering the causality of those features. With the NLP applications are deployed in real world scenario, it is imperative to examined from the causality aspect to unearth the understanding of the model in the alternate scenarios \citep{Moraffah2020, Feder2021}.
\citep{Moraffah2020} has pointed out to the three levels of interpretability listed by \citep{Pearl2018} --- Statistical interpretability, Causal interventional interpretability (Answers the question ``What if") and Causal interpretability (Answers the question ``Why?"). Rest of the survey has focused around the methods related to Statistical interpretability. In this section, this survey will focus on the Causal interpretability. 

Causal interpretability in the NLP are mainly centered around counterfactual explanation (CE) and adversarial examples (AE). There is an ongoing debate around whether CEs and AEs are similar or different. This survey will briefly present that aspect followed by the employed methods in NLP area. 

CEs and AEs are essentially a solution to the same optimization problem equation-\ref{eqn3} 
\begin{equation} \label{eqn3}
\argmax_{x' \in \mathcal{X}} d(x,x') + \lambda d'(f(x'), y_des)
\end{equation}
Where $x$ is the original input and $x'$ is the CE/AE vector, $f$ is the model,$ y_{des}$ is the output, $d$ and $d'$ is the distance, $\lambda$ is the trade-off. parameter \citep{Freiesleben2021}.

As pointed out by \citep{Freiesleben2021}, \citep{Wachter2018} is of opinion that CE and AE are similar in nature but, differing in terms of the objective and in terms of data-points. This view is later countered by \citep{Browne2020} pointing that AE, in all practical scenarios, remain very similar to the real-world input in order to have the imperceptibility. They held the view that they differ in terms of their semantic properties. However, \citep{Verma2021a} held that they are not same because their desiderata are different. Finally, \citep{Freiesleben2021} has tried to put an unified framework for AEs and CEs. Where he differentiated AE and CE on the basis of --- `` In relation to the true instance label and the constraint of how close the respective data point must be". 
\subsubsection{Adversarial examples}
\label{AE}
Evaluation of a model using adversarial examples are more centered towards robustness of the model. By useing AE, one can know the scenario in which its model is going to generate an incorrect output. It will provide an explanation that which type of edit has lead to the change in the output. In order to secure the model from AE attacks, models can be trained on adversarial data. \citep{Ebrahimi2018} has proposed Hot-flip model to generate adversarial examples by flipping the character token. They have further suggested the method to generate AE by word level exchanges. To achieve this they have suggested there must be constraints like similarity, POS preservation etc, so that semantic should not be altered. They have applied in Text classification task.  \citep{Ribeiro2018a} has suggested a method called as Semantically Equivalent Adversarial Rules for AE generation. His method also stressed on the point that the AE should preserve the semantic equivalence. This method is applied in different task like Machine translation, VQA, Sentiment Analysis. \citep{Hossam2020} has trained a white box interpretable substitute model to generate AE. \citep{Sato2018b} has proposed a method to perturb the word embedding to generate AE. Perturbations are guided towards existing word in the word embedding space. It will ensure that the resultant can be easily interpreted at sentence level. They have applied generated examples for Sentiment classification, grammatical error detection, category classification. 

\subsubsection{Counterfactual explanations}
\label{CE}
Similar to AE, a simple approach to explanation would be the generation of CE and compare the response of the model for normal input and counterfactual. Using CE, we can estimate the causal effect \citep{Ross2021a, Gardner2020a}. CE generation can be achieve in two ways --- manually and automatically. Manually writing CE for each of the input would be costly while generating it automatically may produce inconsistent counterfactuals. 

Several authors have proposed a solution to this by altering the representations of the text in place of text itself. \citep{Wu2021} has proposed a framework called Polyjuice to create CE. It is domain agnostic in nature. It takes normal input sentence or masked ([BLANK]) input sentence with control command like negation, quantifier to generate CE. Generation of CE is done by transformer based language model GPT-2 \citep{Radford2019LanguageMA}. \citep{Ross2021} has proposed a CE generator model called MiCE. It is based on T5 \citep{Raffel2020ExploringTL} and uses another form of counterfactual called \textit{Contrastive Explanations}. T5 is fine tuned with input sentence and gold labels for the specified task. During counterfactual generation, it takes masked input and inverted label as an input. The amount of masked token is found by binary search and beam search is employed to keep track of the tokens which has altered the results with highest confidence. \citep{Jacovi2021} has also employed \textit{Contrastive Explanations} for interpretability. They have employed the methods similar to \citep{Elazar2021a}.

\citep{Feder2021a} compute the counterfactual representation by pre-training an additional instance of the language representation model employed by the classifier, with an adversarial component designed to ``forget" the concept of choice, while controlling for confounding concepts. \citep{Elazar2021a} has used the concept of probing to generate CE. Principle aim of this method is to develop a model which can take neural representation as an input and produces an output that devoid of a specific information. They have iteratively trained an auxiliary classifier (as the case with probing) and projecting the representations into their null-space.

\citep{Vig2020, Finlayson2021} has used causal mediation analysis. Mediation analysis relies on measuring the change in an output following a counterfactual intervention in an intermediate variable. It assumes Neural model as a graphical model from input to output with neurons as individual components. 

\section{Observation}
\label{obs}
This survey has broadly classified all the approaches into five broader categories. Further, it has highlighted the simple differences between causality-based and non-causal models. Causal methods, discussed in section -- \ref{CEAE}, has some relevance to input perturbation methods discussed in section -- \ref{inputpertu}. In causal methods, the input features are perturbed to generate a different output from the model. \citep{Rathi2019} has used SHAP(\ref{SHAP}) to produce counterfactual explanations. MiCE and PolyJuice use approach similar to gradient based method (\ref{gradient}) to generate counterfactual examples. 

We also observed that there is a lack of a common quantitative measure to measure interpretability. A simple evaluation approach relies on some form of \textit{decrease in performance} of the model. Human-in-loop evaluation techniques are often employed. As pointed out by \citep{Madsen2021}, following the measures of interpretability \citep{Doshi-Velez2017} there exist some standard measures to measure interpretability. Such measures should be applied to build a unified approach.

Further, we would like to point out that some of the model's explanations are meant for Machine learning practitioners. Compared to those methods, Methods like Natural language explanations(\ref{NLE}), Counterfactual explanations (\ref{CE}) and Adversarial examples (\ref{AE}) produces explanation in a simple language.

\section{Conclusion}
\label{conc}
This survey has presented an overview of interpretability methods from a causal and non-causal perspective. In this survey, we have presented a brief overview of the different approaches and some theoretical discussion around those methods. We have presented the representative paper examples along with the specific NLP tasks they want to highlight. 

\bibliography{anthology,custom,references}
\bibliographystyle{acl_natbib}

\end{document}